\documentclass{easychair}

\usepackage{doc}

 \usepackage{makeidx}

\usepackage{rotating}
\usepackage{pdflscape}

\usepackage{listings}
\usepackage{upgreek}
\usepackage{tikz}
\usepackage[noend]{algpseudocode}
\usepackage{float}
\usepackage{xltabular}
\usepackage{tabularx}
\usepackage{cleveref}
\usepackage{makecell}
\usetikzlibrary{automata, positioning, calc, decorations.pathreplacing}
\usepackage{amsmath}
\usepackage{fontawesome}
\usepackage{adjustbox}
\usepackage{array}
\usepackage{booktabs}
\usepackage{csquotes}
\usepackage{amssymb}
\newtheorem{thm}{Theorem}[section]
\newtheorem{theorem}[thm]{Theorem}
\newtheorem{definition}{Definition}[section]
\newtheorem{lemma}[thm]{Lemma}
\renewcommand{\thefootnote}{\dag}
\usetikzlibrary{arrows.meta}
\usepackage{caption}
\usepackage{xcolor}
\definecolor{keywordcolor}{rgb}{0.7, 0.1, 0.1}   
\definecolor{tacticcolor}{rgb}{0.0, 0.1, 0.6}    
\definecolor{commentcolor}{rgb}{0.4, 0.4, 0.4}   
\definecolor{symbolcolor}{rgb}{0.0, 0.1, 0.6}    
\definecolor{sortcolor}{rgb}{0.1, 0.5, 0.1}      
\definecolor{attributecolor}{rgb}{0.7, 0.1, 0.1} 

\definecolor{academicpurple}{rgb}{0.4,0.0,0.35}
\newcommand{\secref}[1]{\hyperref[#1]{\textcolor{academicpurple}{§\ref*{#1}}}}

\newcommand{\citeref}[1]{\textcolor{academicpurple}{\cite{#1}}}
\newcommand{\defref}[1]{\textcolor{academicpurple}{(\ref{#1})}}

\newcommand{\thmref}[1]{\textcolor{academicpurple}{\hyperref[#1]{Theorem~\ref*{#1}}}}

\newcommand{\eqrefcol}[1]{\textcolor{academicpurple}{\hyperref[#1]{Equation~(\ref*{#1})}}}

\newcommand{\extlink}{~\ensuremath{{}^\text{\faExternalLink}}}

\makeatletter
\def\do@url@hyp{\do\-}
\makeatother

\definecolor{codebg}{rgb}{0.96,0.96,0.99}
\definecolor{codeframe}{rgb}{0.75,0.70,0.80}
\title{Formalized Hopfield Networks and Boltzmann Machines}

\author{
Matteo Cipollina\inst{1}\textsuperscript{\dag}
\and
    Michail Karatarakis\inst{2}\textsuperscript{\dag}
\and
   Freek Wiedijk\inst{2}
}

\institute{
  Axiomatic AI,
  Barcelona, Spain\\
  \email{matteo.cipollina@yahoo.it}
\and
   Radboud University, Nijmegen, The Netherlands\\
   \email{michail.karatarakis@ru.nl, freek@cs.ru.nl}\\
 }

\authorrunning{Cipollina, Karatarakis and Wiedijk}

\titlerunning{Formalized Hopfield Networks and Boltzmann Machines}

\begin{document}

\maketitle
\footnotetext{Equal contribution, authors listed alphabetically.} \renewcommand{\thefootnote}{\arabic{footnote}} \setcounter{footnote}{0}
\begin{abstract}
Neural networks are widely used, yet their analysis and verification remain challenging. We present
a Lean~4 formalization covering both deterministic and stochastic models. We first formalize
Hopfield networks -- recurrent networks that store patterns as stable states -- and prove their
convergence, and the correctness of Hebbian learning, the rule that updates parameters to encode
patterns. We then turn to stochastic networks, whose probabilistic updates converge to a stationary
distribution: we formalize the dynamics and learning of Boltzmann machines and prove their
ergodicity -- convergence to a \emph{unique} stationary distribution -- via a new formalization of the Perron--Frobenius theorem.

\end{abstract}

\setcounter{tocdepth}{2}

\section{Introduction}\label{sec:1}
Neural networks underpin most current practice in machine learning. In October 2024, John Hopfield and Geoffrey Hinton were awarded the Nobel Prize in Physics for their work in machine learning, particularly in the development of artificial neural networks~\citeref{nobel}. The prize recognizes models whose mathematics comes directly from statistical mechanics. 

One of Hopfield’s most influential achievements is the Hopfield network~\citeref{hopfield1982neural}, introduced in 1982. Hopfield networks model associative memory, storing and recalling patterns from partial or noisy input, and were originally inspired by the Ising model of magnetism~\citeref{Ising:1925em}. A key feature is their convergence property, where the network reaches a stable state corresponding to a stored pattern. Building on Hopfield networks, Boltzmann machines were introduced by Ackley, Hinton and Sejnowski~\citeref{hinton} in 1985 as a stochastic generalization. In Boltzmann machines, neuron updates are probabilistic rather than deterministic, allowing the network to sample from probability distributions over states instead of converging to a single stable state.

In 2020, Ramsauer et al.\ showed that the update rule of continuous Hopfield networks is equivalent to the attention mechanism used in Transformers~\citeref{modern}. These models also have a concrete computational role: minimizing a Hopfield energy is, up to an
affine change of variables, quadratic unconstrained binary optimization
(QUBO)~\citeref{lucas2014ising, glover2022quantum}, the native input format of quantum annealers. The same energy descent is realized directly in analog neuromorphic
hardware~\citeref{guo2015modeling, li2020cmos}, while stochastic Ising machines realize Boltzmann sampling over discrete spin states~\citeref{camsari2017pbits, niazi2024boltzmann} -- optimization and sampling carried out in the physical device rather than a general-purpose processor. A
machine-checked account of these dynamics therefore gives a specification against which such hardware and the algorithms running on it could be verified. 

Mainstream neural-network verification targets feedforward networks, where the network is a fixed function and the properties are local: typically robustness of a single input-output map on a bounded region \citeref{Botoeva_Kouvaros_Kronqvist_Lomuscio_Misener_2020, NEURIPS2018_be53d253, imandra, 10.1007/978-3-319-68167-2_19,10.1007/978-3-030-53288-8_2,8953865,OPT-035}. In contrast, the formalization of recurrent networks and their learning algorithms themselves remains relatively unexplored. Energy-based recurrent models demand global statements of a different kind: that asynchronous updates terminate, that the induced Markov chain is ergodic, and that it converges to a unique stationary distribution -- theorems of probability theory and statistical mechanics, not of bound propagation. 

This paper addresses that gap, and the project was inspired by a suggestion from Britt Anderson in the Lean Zulip chat~\citeref{britt}. We formalize both deterministic Hopfield networks and stochastic Boltzmann machines, giving a rigorous foundation for reasoning about their dynamics and learning. Engaging with these models also led to a general framework treating feedforward and recurrent architectures uniformly.

\noindent\textbf{Contributions and overview.}
To our knowledge, this is the first formalization of Hopfield networks and Boltzmann machines in any proof assistant. The development comprises around 16{,}000 lines of new Lean code (excluding blank lines and comments), of which roughly 7{,}100 are reusable mathematical infrastructure (probability and Perron--Frobenius) and 9{,}000 the neural-network development; \Cref{tab:contributions} gives the per-component split. The proofs use standard Lean and \texttt{mathlib} tactics and automation throughout.
We work in Lean~4 because \texttt{mathlib} offers a single unified hierarchy of probability, algebra, analysis, and topology, all used together in the proofs below; aiming to contribute back, we follow its philosophy of maximal generality rather than ad hoc formalization. We also bridge to the statistical-mechanics part of \texttt{Physlib}, a \texttt{mathlib}-style physics library, and lay groundwork for \texttt{Optlib}~\citeref{optlib2025} (a library of first-order optimization and convergence theorems). Every component in \Cref{tab:contributions} is new; the table records its size, computability and ties to existing Lean libraries. Most definitions in \secref{sec:3} -- \secref{sec:5} are \lstinline{noncomputable}, relying on \texttt{mathlib}'s nonconstructive reals, so they yield no executable code. The clickable symbol {\faExternalLink} links to the corresponding source, and a public repository accompanies the paper\,\href{https://github.com/mkaratarakis/HNBM}{\extlink}. 
We conclude with related work (\secref{sec:6}) and future directions (\secref{sec:7}).

\begingroup
\footnotesize
\setlength{\tabcolsep}{4pt}
\renewcommand{\arraystretch}{1.2}
\begin{xltabular}{\textwidth}{@{}>{\raggedright\arraybackslash}X r c >{\raggedright\arraybackslash}p{0.20\textwidth} >{\raggedright\arraybackslash}p{0.15\textwidth}@{}}
\toprule
Component & Lines & Comp. & Reuse & Upstreaming \\
\midrule
Unified framework for feedforward and recurrent networks (\secref{sec:2}). The organizing
\texttt{Quiver} core was \emph{discovered}, not designed: adopted for path manipulation in the
Perron--Frobenius proof, it proved the right abstraction for the whole framework.
  & 2{,}400 & yes
  & \texttt{mathlib}: \texttt{Quiver},\newline linear algebra, analysis
  & \texttt{Physlib}\newline (PR under review) \\
Discrete symmetric Hopfield networks: convergence and Hebbian learning (\secref{sec:2.4}).
Formalization surfaced an implicit \emph{fairness} condition on the update order, required for
convergence yet omitted from textbook statements (\thmref{hnfairconvergence}).
  & 2{,}000 & yes
  & \texttt{mathlib}: \texttt{Matrix}, \texttt{Vector}
  & \texttt{Physlib}\newline (PR under review) \\
Probabilistic infrastructure: reversibility, invariance of Markov kernels, Gibbs sampling (\secref{sec:3}).
  & 1{,}700 & no
  & \texttt{mathlib}: \texttt{Kernel}, \texttt{PMF},\newline linear algebra
  & \texttt{mathlib}\newline (ongoing) \\
Perron--Frobenius for irreducible/stochastic matrices (\secref{sec:4}); first in Lean~4, via an analytic Collatz--Wielandt argument.
  & 5{,}400 & no
  & \texttt{mathlib}: spectrum, topology,\newline \texttt{Quiver} paths
  & \texttt{mathlib}\newline (PR in prep.) \\
Boltzmann machines: convergence to the Boltzmann distribution, plus single-pattern Boltzmann learning and CD-$k$ (\secref{sec:5}, \secref{sec:5.2}).
  & 4{,}600 & no
  & \texttt{Physlib}: \texttt{CanonicalEnsemble}, \texttt{Temperature};\newline \texttt{mathlib}: probability, analysis
  & \texttt{Physlib}\newline (PR in prep.);\newline \texttt{Optlib}\newline (submission in prep.) \\
\bottomrule
\end{xltabular}
\addtocounter{table}{-1}
\captionof{table}{Contributions. \enquote{Comp.} records whether the component yields executable code; \enquote{Reuse} lists the existing library components it builds on; \enquote{Upstreaming} gives the status of contributing it back.} \label{tab:contributions}
\endgroup

\section{General Neural Networks}\label{sec:2}
A fundamental design decision in formalizing neural networks is the choice of the underlying representation. Neural networks are commonly represented as directed graphs, categorized as feedforward (acyclic) or recurrent (cyclic) and typically treated as distinct mathematical objects in the literature (e.g., Ch.~5 vs. Ch.~8, \citeref{comp}). Modern deep learning frameworks, such as TensorFlow \citeref{tensorflow} and PyTorch \citeref{pytorch}, predominantly adopt a sequential model, in which networks are represented as a sequence of layers -- typically a list paired with an abstract table of activation functions -- or, more generally, as a directed acyclic graph (DAG) of tensor operations, i.e., a computational graph. This sequential approach is also reflected in the verification literature (see \secref{sec:6}). In contrast, the definition we give below is general enough to accommodate both architectures and integrate with \texttt{Physlib} (\secref{sec:5.1}). Feedforward networks arise as a special case (see \texttt{SequentialArch}\href{https://github.com/mkaratarakis/HNBM/blob/a9a7755008a756ca416b304100e44d52bc203ef6/HopfieldNet/Quiver/NeuralNetwork/Seq.lean#L17}{\extlink}, \lstinline{Sequential.toNeuralNetwork}\href{https://github.com/mkaratarakis/HNBM/blob/a9a7755008a756ca416b304100e44d52bc203ef6/HopfieldNet/Quiver/NeuralNetwork/Seq.lean#L56}{\extlink}). This
work is guided by the theory in Chapter~4 of \citeref{comp} and builds on \texttt{mathlib}'s graph
theory.

Let $G = (U, C)$ be a directed graph consisting of a finite set $U$ of vertices (or nodes) and a finite set $C \subseteq U \times U$ of edges. An edge $e = (u, v) \in C$ is said to be \emph{directed} from the vertex $u$ to the vertex $v$. The predecessors and successors of a vertex \( u \) in a directed graph \( G = (U, C) \) are defined as
$\mathrm{pred}(u) = \{ v \in U \mid (v, u) \in C \}$ and $\mathrm{succ}(u) = \{ v \in U \mid (u, v) \in C \}$ respectively.

\begin{definition}[Def. 4.3, \citeref{comp}]\label{nndef}
An (artificial) \emph{neural network} is a directed graph \( G = (U, C) \), where neurons \( u \in U \) are connected by directed edges \( c \in C \) (connections). The neuron set is partitioned as \( U = U_{\mathrm{in}} \cup U_{\mathrm{out}} \cup U_{\mathrm{hidden}} \), with \( U_{\mathrm{in}}, U_{\mathrm{out}} \neq \emptyset \) and \( U_{\mathrm{hidden}} \cap (U_{\mathrm{in}} \cup U_{\mathrm{out}}) = \emptyset \). Each connection \( (v, u) \in C \) has a weight \( w_{uv} \), and each neuron \( u \) has real-valued quantities: network input \( \mathrm{net}_u \), activation \( \mathrm{act}_u \), and output \( \mathrm{out}_u \). Input neurons \( u \in U_{\mathrm{in}} \) also have a fourth quantity, the external input \( \mathrm{ext}_u \). Each neuron \( u \) is associated with the following functions:
$$f_{\mathrm{net}}^{(u)} : \mathbb{R}^{2|\mathrm{pred}(u)|+ \kappa_1 (u)} \to \mathbb{R}, \quad
 f_{\mathrm{act}}^{(u)} : \mathbb{R}^{1+\kappa_2 (u)} \to \mathbb{R} \quad \mbox{and} \quad f_{\mathrm{out}}^{(u)} : \mathbb{R} \to \mathbb{R}. $$
These functions compute \( \mathrm{net}_u \), \( \mathrm{act}_u \), and \( \mathrm{out}_u \), where \( \kappa_1(u) \) and \( \kappa_2(u) \) count the number of parameters of those functions, which can depend on the neurons. 
\end{definition}

\begin{figure}[H]
\vspace{-10pt}
    \centering
    \begin{minipage}[b]{0.43\textwidth}
        \centering
        \begin{tikzpicture}[
            neuron/.style={circle, draw, minimum size=0.8cm, thick, fill=blue!5},
            connect/.style={-Stealth, thick},
            weight/.style={font=\small, color=red!70!black},
            scale=0.70, every node/.style={transform shape}
        ]
            \node[neuron] (u1) at (0, 2.5) {$u_1$};
            \node[neuron] (u2) at (0, 0) {$u_2$};
            
            \node[neuron] (u3) at (3, 2.5) {$u_3$};
            \node[neuron] (u4) at (3, 0) {$u_4$};
            
            \node[neuron] (u5) at (6, 1.25) {$u_5$};

            \draw[connect] (-1.5, 2.5) -- (u1) node[midway, above] {$x_1$};
            \draw[connect] (-1.5, 0) -- (u2) node[midway, above] {$x_2$};
            
            \draw[connect] (u5) -- (7.5, 1.25) node[midway, above] {$y$};

            \draw[connect] (u1) -- (u3) node[pos=0.4, above, weight] {0.5};
            \draw[connect] (u1) -- (u4) node[pos=0.3, above, weight, sloped] {$-1$};
            \draw[connect] (u2) -- (u3) node[pos=0.3, below, weight, sloped] {1.2};
            \draw[connect] (u2) -- (u4) node[pos=0.4, below, weight] {0.8};

            \draw[connect] (u3) -- (u5) node[pos=0.4, above, weight, sloped] {1.5};
            \draw[connect] (u4) -- (u5) node[pos=0.4, below, weight, sloped] {$-2$};

        \end{tikzpicture}
    \end{minipage}\hfill
    \begin{minipage}[b]{0.48\textwidth}
        \centering
        \begin{tikzpicture}[
            neuron/.style={circle, draw, minimum size=0.8cm, thick, fill=blue!5},
            connect/.style={-Stealth, thick},
            weight/.style={font=\small, color=red!70!black},
            scale=0.70, every node/.style={transform shape} 
        ]
            \node[neuron] (u1) at (0, 2.5) {$u_1$};
            \node[neuron] (u2) at (0, 0) {$u_2$};
            \node[neuron] (u3) at (2.8, 1.25) {$u_3$};

            \draw[connect] (-1.2, 2.5) -- (u1) node[midway, above] {$x_1$};
            \draw[connect] (-1.2, 0) -- (u2) node[midway, above] {$x_2$};
            \draw[connect] (u3) -- (4.0, 1.25) node[midway, above] {$y$};

            \draw[connect] (u1) -- (u2) node[midway, left, weight] {1};
            \draw[connect] (u1) -- (u3) node[pos=0.4, above, weight] {$-2$};
            \draw[connect] (u2) -- (u3) node[pos=0.4, below, weight] {3};
            
            \draw[connect] (u3) to[bend right=50] node[midway, above, weight] {4} (u1);
        \end{tikzpicture}
    \end{minipage}
    
    \vspace{0.1cm} 
    \begin{minipage}[t]{0.48\textwidth}
\caption{A feedforward network with external inputs $x_1,x_2$ and output $y$, and no loops; arrow labels are connection weights.}
        \label{fig:strict_feedforward}
    \end{minipage}\hfill
    \begin{minipage}[t]{0.48\textwidth}
                \captionof{figure}{A recurrent network, same conventions as Fig.~\ref{fig:strict_feedforward}. The feedback edge $u_3 \to u_1$ lets information cycle back.}
        \label{fig:recurrent_network} 
    \end{minipage}
    \vspace{-15pt}
\end{figure}
In the index of a weight \( w_{uv} \), the neuron receiving the connection is listed first, following the \enquote{row first, then column} convention of the adjacency matrix, with matrix-vector equations written using column vectors and standard matrix-vector multiplication rules. This structure groups all incoming weights to a neuron within the same row, represented compactly as the matrix $W = (w_{uv})_{u,v \in U}.$ The new activation $\mathrm{act}_u'$ of a neuron $u$ is computed as follows:
$$
\mathrm{act}_u'=  
f_{\mathrm{act}}^{(u)} \big(f_{\mathrm{net}}^{(u)} \big(
w_{uv_1}, \ldots, w_{uv_{\mathrm{pred}(u)}}, f_{\mathrm{out}}^{(v_1)}(\mathrm{act}_{v_1}),\ldots,
f_{\mathrm{out}}^{(v_{\mathrm{pred}(u)})}(\mathrm{act}_{v_{\mathrm{pred}(u)}}),
\boldsymbol{\sigma}^{(u)}\big), \boldsymbol{\theta}^{(u)}\big)
$$ where $\boldsymbol{\sigma}^{(u)} = (\sigma_1^{(u)} , \ldots , \sigma_{\kappa_1(u)}^{(u)} )$ and $\boldsymbol{\theta}^{(u)} = (\theta_1^{(u)} , \ldots , \theta_{\kappa_2(u)}^{(u)} )$ are the input parameter vectors.

The following diagram shows the structure of a generalized neuron abstracting away specific model assumptions. Here $\mathbf{act} = (\mathrm{act}_{v_1}, \ldots , \mathrm{act}_{v_{|U|}})$ and $\mathbf{out} = (\mathrm{out}_{v_1}, \ldots, \mathrm{out}_{v_{|U|}})$ collect the activations and outputs of all neurons, and $v_i$ ranges over the predecessors of $u$.
\begin{figure}[H]

\centering
\scalebox{0.85}{%
\begin{tikzpicture}[node distance=2cm]
    \node (fout) [draw, minimum width=2cm, minimum height=1cm] {$f_{\mathrm{out}}^{(v_i)}$};
    \node (act) [left of=fout] {$\boldsymbol{act_{v_i}}$};
    \node (out) [right of=fout] {$\boldsymbol{out_{v_i}}$};
    \node (fnet) [right of=out, draw, minimum width=2cm, minimum height=1cm] {$f_{\mathrm{net}}^{(u)}$};
    \node (net) [right of=fnet] {$\mathrm{net}_{u}$};
    \node (fact) [right of=net, draw, minimum width=2cm, minimum height=1cm] {$f_{\mathrm{act}}^{(u)}$};
    \node (actu') [right of=fact] {${\mathrm{act}}_{u}'$};
    \draw[->] (act) -- (fout);
    \draw[->] (fout) -- (out);
    \draw[->] (out) -- (fnet);
    \draw[->] (fnet) -- (net);
    \draw[->] (net) -- (fact);
    \draw[->] (fact) -- (actu');
    \node (sigma) [above of=fnet] {$\boldsymbol{w}$};
    \node (w) [above right of=fnet] {$\boldsymbol{\sigma}$};
    \node (theta) [above right of=fact] {$\boldsymbol{\theta}$};
    \draw[->] (sigma) -- (fnet);
    \draw[->] (w) -- (fnet);
    \draw[->] (theta) -- (fact);
\end{tikzpicture}%
}
\caption{Structure of a generalized neuron.}
\label{fig:generalized_neuron}
\end{figure}

In our formalization, we separate a neural network into its architecture, defined by its graph structure \lstinline{NeuralNetwork}\href{https://github.com/mkaratarakis/HNBM/blob/a9a7755008a756ca416b304100e44d52bc203ef6/HopfieldNet/Quiver/NeuralNetwork/Main.lean#L16}{\extlink} 
(see Listing \ref{lst:neural_def}), and parameters 
\lstinline{Params}\href{https://github.com/mkaratarakis/HNBM/blob/a9a7755008a756ca416b304100e44d52bc203ef6/HopfieldNet/Quiver/NeuralNetwork/Main.lean#L46}{\extlink} 
(see Listing \ref{lst:params_def}), controlling dynamics like input processing. This separation allows flexibility, enabling different functions, weights, or learning rules without redefining the architecture. For brevity, we omit well-formedness conditions, which are fully specified in the repository. 

\begin{lstlisting}[label={lst:neural_def},  stepnumber=1, numbersep=5pt, aboveskip=3pt, belowskip=3pt,captionpos=b, caption={Definition of neural network structure.}]
universe uR uU uζ v
structure NeuralNetwork (R : Type uR) (U : Type uU) (ζ : Type uζ) [Zero R] 
  extends Quiver.{v, uU} U where
  (Ui Uo Uh : Set U)
  (hUi : Ui ≠ ∅)
  (hUo : Uo ≠ ∅)
  (hU : Set.univ = (Ui ∪ Uo ∪ Uh))
  (hhio : Uh ∩ (Ui ∪ Uo) = ∅)
  (κ1 κ2 : U → ℕ)
  (fnet : ∀ u : U, (U → R) → (U → R) → Vector R (κ1 u) → R)
  (fact : ∀ u : U, ζ → R → Vector R (κ2 u) → ζ)
  (fout : ∀ _ : U, ζ → R)
  (m : ζ → R)
  (pact : ζ → Prop)
  (pw : ∀ (u v : U), (u ⟶ v) → Prop)
  (pwMat : Matrix U U Prop := fun u v => ∃ f : (u ⟶ v), pw u v f)
  (pm : Matrix U U R → Prop)
\end{lstlisting}

\begin{lstlisting}[label={lst:params_def},  stepnumber=1, numbersep=5pt, aboveskip=3pt, belowskip=3pt,captionpos=b, caption = Definition of parameters.]
variable {R U ζ : Type} [Zero R]
structure Params (NN : NeuralNetwork R U ζ) where
  (h_arrows : ∀ u v (f : NN.Hom u v), NN.pw u v f)
  (w : Matrix U U R) -- the weight matrix W
  (σ : ∀ u : U, Vector R (NN.κ1 u)) -- the input vector
  (θ : ∀ u : U, Vector R (NN.κ2 u)) -- the threshold vector
  (hw : ∀ u v, ¬ NN.pwMat u v → w u v = 0)
  (hw' : NN.pm w)
\end{lstlisting}
To ensure flexibility and generality across network structures, we used an arbitrary type \lstinline{R} for weights, with \lstinline{[Zero R]} indicating the existence of a zero element in \lstinline{R}, instead of restricting ourselves to real numbers. Similarly, we use \lstinline{U} for the neurons. However, standard definitions of neural networks model activations simply as real numbers, giving them the same type as the weights. Since activations may be discrete whereas weights may be any real number, we instead parameterize the network over a separate abstract activation type \lstinline{ζ}, using the embedding \lstinline{(m : ζ → R)} to interpret these states as elements of the weight type. This design unifies architectures via the \lstinline{TwoStateNeuralNetwork}\label{twostate}\href{https://github.com/mkaratarakis/HNBM/blob/a9a7755008a756ca416b304100e44d52bc203ef6/HopfieldNet/Quiver/NeuralNetwork/TwoState.lean#L205}{\extlink} 
typeclass, which abstracts over concrete encodings such as $\{-1,+1\},$ $\{0,1\},$ or custom types like \lstinline{Signum} and ensures compatibility with the \texttt{Physlib} library (see \secref{sec:5.1}). The construction is also parameterized by an interpreter \lstinline{(actMap : ζ → R → R)}, which assigns a concrete function \lstinline{R → R} to each abstract descriptor. This decoupled design allows the architecture to remain abstract while supporting heterogeneous activations -- such as ReLU for hidden layers and sigmoid for output -- within a single network instance.

\noindent\textbf{Design decisions}
We initially modeled neural networks using \texttt{mathlib}'s \texttt{Digraph} structure, which defines a graph as a binary relation on vertices :
\begin{lstlisting}[numberstyle=\tiny, stepnumber=1, numbersep=5pt, aboveskip=3pt, belowskip=3pt]
structure Digraph (V : Type*) where Adj : V → V → Prop
\end{lstlisting} In the textbook definition~\ref{nndef}, function arguments and summations for a neuron $u$ are generally indexed running over the set $\mathrm{pred}(u)$. In Lean that could correspond to passing around adjacency proof objects everywhere. For example, the weight matrix could have been written as \lstinline{(w : ∀ u v : U, Adj v u → R)}. Similarly, the type of \( f_{\mathrm{net}}^{(u)} : \mathbb{R}^{2|\mathrm{pred}(u)| + \kappa_1 (u)} \to \mathbb{R} \) in Lean would have been :
\begin{lstlisting}[stepnumber=1, numbersep=5pt, aboveskip=3pt, belowskip=3pt]
(fnet : ∀ u : U, (∀ v : U, Adj v u → R) → (∀ v : U, Adj v u → R) → Vector R (κ1 u) → R)
\end{lstlisting}
This then explicitly needs tracking adjacency relations, complicating everything. To simplify our formalization, we decided not to do this. Instead we used the convention that all weights for neurons that are not adjacent are zero. Therefore we used a matrix \lstinline{(w : Matrix U U R)} directly and the function formalizing \( f_{\mathrm{net}}^{(u)} \) had type 
\begin{lstlisting}[numberstyle=\tiny, stepnumber=1, numbersep=5pt, aboveskip=3pt, belowskip=3pt]
(fnet : ∀ u : U, (U → R) → (U → R) → Vector R (κ1 u) → R)
\end{lstlisting}
This may be less efficient for sparse networks needing precise connection control, and can make the functions harder to implement. Adjacency could be supplied as extra \lstinline{(U → Bool)} arguments, but our applications did not need this, so it is not part of our formalization. Furthermore, whilst a propositional representation suffices for simple connectivity, the formalization of the Perron--Frobenius theorem in~\secref{sec:4} necessitated a transition to \texttt{mathlib}'s \texttt{Quiver} \href{https://github.com/leanprover-community/mathlib4/blob/8f9d9cff6bd728b17a24e163c9402775d9e6a365/Mathlib/Combinatorics/Quiver/Basic.lean#L38}{\extlink} 
type. A quiver on a vertex type \lstinline{V} assigns a type \lstinline{a ⟶ b} of directed arrows to every pair of vertices \lstinline{(a b : V)}.
\begin{lstlisting}[language=Lean]
class Quiver (V : Type u) where Hom : V → V → Type v
\end{lstlisting}
This generalizes \lstinline{Digraph V}, where the arrow \lstinline{Hom : a ⟶ b} is a proposition and not a type. By adopting \lstinline{Quiver} as our foundation, connections inhabit a \lstinline{Type}, enabling the representation of multigraphs with multiple distinct edges between vertices. Furthermore, \texttt{Quiver} benefits from a significantly more mature API in \texttt{mathlib} and provides the extensibility required to model complex neural architectures beyond those explored in this work.

\noindent\textbf{Operation of neural networks.}
Computation proceeds in two phases. In the \emph{input phase}, each input neuron is set to its external input and all other neurons to $0$, and the output function is applied to these activations so that every
neuron produces an output. In the \emph{work phase}, neurons repeatedly recompute their activations -- and hence their outputs -- in an order that is not fixed: either simultaneously
(synchronous update), using the previous outputs, or one at a time (asynchronous update), using the most recently computed ones. Different update schemes suit different network types.

Each neuron is either \emph{firing} or \emph{not firing}; the two states are encoded numerically and enforced by an admissibility predicate on activations. The general convention
uses $1$ (firing) and $0$ (not firing), so a state is a vector
$s = (\mathrm{act}_{u_1}, \ldots, \mathrm{act}_{u_n})^\top \in \{0,1\}^n$, whereas Hopfield
networks use $\pm 1$ ($\mathrm{act}_u \in \{-1,1\}$). The \texttt{State}\href{https://github.com/mkaratarakis/HNBM/blob/a9a7755008a756ca416b304100e44d52bc203ef6/HopfieldNet/Quiver/NeuralNetwork/Main.lean#L62}{\extlink} structure stores the
activations \lstinline{(act : U → ζ)} together with a proof that each is admissible. Functions
\lstinline{State.fout} and \lstinline{State.fnet} operate on these stored activations, playing
the role of \lstinline{NeuralNetwork.fout} and \lstinline{NeuralNetwork.fnet} for the current
state, while \lstinline{State.Up}\href{https://github.com/mkaratarakis/HNBM/blob/a9a7755008a756ca416b304100e44d52bc203ef6/HopfieldNet/Quiver/NeuralNetwork/Main.lean#L87}{\extlink} updates a single chosen neuron \texttt{u} via
\lstinline{NeuralNetwork.fact}, leaving all other activations unchanged.

In practice, a stopping criterion -- based on updates, energy, or convergence measures -- ends the iteration. Stability is the property \lstinline{State.isStable}\href{https://github.com/mkaratarakis/HNBM/blob/a9a7755008a756ca416b304100e44d52bc203ef6/HopfieldNet/Quiver/NeuralNetwork/Main.lean#L107}{\extlink}: a state \lstinline{s} is
stable if applying \lstinline{State.Up} to any neuron leaves that neuron's activation
unchanged. Given an update sequence \lstinline{(useq : ℕ → U)}, \lstinline{seqStates}\href{https://github.com/mkaratarakis/HNBM/blob/a9a7755008a756ca416b304100e44d52bc203ef6/HopfieldNet/Quiver/NeuralNetwork/Main.lean#L103}{\extlink} constructs
the resulting sequence of states recursively -- the first is the initial state \lstinline{s}, and each subsequent one updates the neuron \lstinline{useq} names at that step. The \lstinline{workPhase}\href{https://github.com/mkaratarakis/HNBM/blob/a9a7755008a756ca416b304100e44d52bc203ef6/HopfieldNet/Quiver/NeuralNetwork/Main.lean#L100}{\extlink} function instead takes an initial state \lstinline{extu} and a finite list \lstinline{uOrder}, folding \lstinline{State.Up} over it (\lstinline{List.foldl}) to perform one update per list entry, in order, and return the resulting state.

As a sanity check we validate the framework on the three-neuron recurrent network of
Fig.~\ref{fig:recurrent_network} (p.~42, \citeref{comp},\href{https://github.com/mkaratarakis/HNBM/blob/a9a7755008a756ca416b304100e44d52bc203ef6/HopfieldNet/Quiver/NeuralNetwork/test.lean#L15}{\extlink})
: under one update order the work phase reaches a stable all-zero state, while another order oscillates indefinitely, so the output can
depend not only on the input but also on the number of updates -- behavior precluded in Hopfield networks, which always converge under asynchronous updates.

\section{Hopfield networks}\label{sec:2.4}
Before Hopfield's work, neural networks were primarily feedforward models, with unidirectional information flow. Hopfield networks popularized recurrent connections, enabling feedback between neurons, allowing them to model dynamic systems and handle more complex tasks. We follow the mathematical background in Chapter 8 of \citeref{comp}.
\begin{definition}
A \emph{Hopfield network} is a neural network with graph $G = (U,C)$ as described in the previous section, that satisfies the following conditions:
\( U_{\mathrm{hidden}} = \emptyset \), and \( U_{\mathrm{in}} = U_{\mathrm{out}} = U \), \( C = U \times U - \{(u, u) \mid u \in U \} \), i.e., no self-connections. The connection weights are symmetric, i.e., \( \forall u, v \in U \), we have \( w_{uv} = w_{vu} \) when \( u \neq v \). The activation of each neuron is either \( 1 \) or \( -1 \) depending on the input. There are no loops, i.e., no neuron receives its own output as input: $w_{uu} = 0$ for all $u \in U$. \noindent The neuron functions for the Hopfield network are defined for all \( u \in U \) as:
$$
\begin{aligned}
f^{(u)}_{\mathrm{net}}(w_u,\mathrm{in}_u)
&= \!\sum_{v \in U \setminus \{u\}}\! w_{uv}\,\mathrm{out}_v, \label{fnet}\;\enskip
f^{(u)}_{\mathrm{act}}(\mathrm{net}_u,\theta_u)
= \begin{cases}
1 & \text{if } \mathrm{net}_u \ge \theta_u,\\
-1 & \text{otherwise},
\end{cases}\label{fact}\;
f^{(u)}_{\mathrm{out}}(\mathrm{act}_u)=\mathrm{act}_u \label{fout}.
\end{aligned}
$$
\end{definition}

Here $\theta_u$ is the firing threshold: increasing $\theta_u$ makes activation more difficult. This is consistent with the energy convention below, where the threshold contribution is $+\theta_u\,\mathrm{act}_u$.

We define the \lstinline{HopfieldNetwork}
\href{https://github.com/mkaratarakis/HNBM/blob/a9a7755008a756ca416b304100e44d52bc203ef6/HopfieldNet/Quiver/HN/Core.lean#L50}{\extlink}
as an instance of \lstinline{NeuralNetwork} over a finite type \lstinline{U} and a strictly ordered ring \lstinline{R}. The network topology is fully connected excluding self-loops, formalized by defining the edge type \lstinline{Hom u v} as \lstinline{PLift (u ≠ v)}. The sets of input \lstinline{Ui} and output \lstinline{Uo} neurons coincide with the universal set \lstinline{Set.univ}, while the set of hidden neurons \lstinline{Uh} is empty. 

As the three-neuron example of \secref{sec:2} \href{https://github.com/mkaratarakis/HNBM/blob/a9a7755008a756ca416b304100e44d52bc203ef6/HopfieldNet/Quiver/NeuralNetwork/test.lean#L15}{\extlink} shows, updates can cause a recurrent network to oscillate between activation states. However, asynchronous updates (one neuron at a time) in a Hopfield network always reach a stable state, preventing oscillation.

\begin{thm}[Convergence Theorem for Hopfield networks: Theorem 8.1, \citeref{comp}]\label{hnfairconvergence}
If the activations of the neurons of a Hopfield network are updated asynchronously, a stable state is reached in a finite number of steps (\lstinline{HopfieldNet_convergence_fair})\href{https://github.com/mkaratarakis/HNBM/blob/a9a7755008a756ca416b304100e44d52bc203ef6/HopfieldNet/Quiver/HN/Core.lean#L725}{\extlink}.
If the neurons are traversed in an arbitrary, but fixed cyclic fashion, at most $n\cdot2^n$ steps (updates of individual neurons) are needed, where $n$ is the number of neurons of the network (\lstinline{HopfieldNet_convergence_cyclic})\href{https://github.com/mkaratarakis/HNBM/blob/a9a7755008a756ca416b304100e44d52bc203ef6/HopfieldNet/Quiver/HN/Core.lean#L825}{\extlink}.
\end{thm}

This theorem is proved by defining an energy function that assigns a real value to each state of the Hopfield network, which decreases or remains constant with each transition. When energy remains unchanged, we show that a state can never be revisited. Since there are only finitely many states, this ensures the network will eventually reach a stable state. 

The energy function \lstinline{E} of a Hopfield network with \( n \) neurons \( u_1, \ldots, u_n \) is
\begin{equation}\label{energy}
E = -\frac{1}{2} \sum_{\substack{u,v \in U \\ u \neq v}} w_{uv} \, act_u \, act_v + \sum_{u \in U} \theta_u \, act_u \href{https://github.com/mkaratarakis/HNBM/blob/a9a7755008a756ca416b304100e44d52bc203ef6/HopfieldNet/Quiver/HN/Core.lean#L243}{\extlink}.
\end{equation} 
The factor \( \frac{1}{2} \) compensates for the symmetry of the weights, as each term in the first sum appears twice. We show that energy cannot increase during a state transition when a neuron is updated. In an asynchronous update, only one neuron's activation, \( u \), changes from \( \text{act}_u^{(\text{old})} \) to \( \text{act}_u^{(\text{new})} \). The energy difference between the old and new states consists only of terms containing \( \text{act}_u \), with all other terms canceling out. The factor \( \frac{1}{2} \) vanishes due to the symmetry of the weights, causing each term to occur twice. Then we can extract the new and old activation states of neuron \( u \) from the sums, yielding the energy difference (\lstinline{energy_diff_formula}):

$$\Delta E = E^{(\text{new})} - E^{(\text{old})}
        = (\text{act}_u^{(\text{old})} - \text{act}_u^{(\text{new})}) 
\bigg( \underbrace{\sum_{v \in U - \{u\}} w_{uv} \, \text{act}_v}_{\mathrm{net}_u} - \theta_u\bigg)\href{https://github.com/mkaratarakis/HNBM/blob/a9a7755008a756ca416b304100e44d52bc203ef6/HopfieldNet/Quiver/HN/Core.lean#L355}{\extlink}.$$

We consider two cases. If \( \mathrm{net}_u < \theta_u \) and the activation changed, then \( \text{act}_u^{(\text{new})} = -1 \) and \( \text{act}_u^{(\text{old})} = 1 \), so \( \Delta E < 0 \). If \( \mathrm{net}_u \geq \theta_u \), then \( \text{act}_u^{(\text{old})} = -1 \) and \( \text{act}_u^{(\text{new})} = 1 \), yielding \( \Delta E \leq 0 \).

If a state transition reduces energy, the original state cannot be revisited, as this would require an energy increase. In the second case, where energy remains constant, the transition leads to more $+1$ activations, as the updated neuron’s activation changes from $-1$ to $1.$ Thus, the original state cannot be revisited in this case either. The theorem \texttt{energy\_lt\_zero\_or\_pluses\_increase}\href{https://github.com/mkaratarakis/HNBM/blob/a9a7755008a756ca416b304100e44d52bc203ef6/HopfieldNet/Quiver/HN/Core.lean#L417}{\extlink} captures this idea.

\noindent\textbf{Formalization insight}
Each state transition reduces the number of reachable states. Since the total number of states is finite, the network will eventually reach a stable state. However, this convergence is only guaranteed if every neuron is eventually updated. If some neurons are excluded, the network may fail to stabilize. This highlights a logical gap in the standard literature, where \thmref{hnfairconvergence} frequently omits the necessary hypothesis of \enquote{fairness}. Without this constraint, an update schedule may permanently exclude neurons that need to change, causing the energy to stop decreasing even though the network has not yet reached a stable state. To prevent this, we define \enquote{fairness}\href{https://github.com/mkaratarakis/HNBM/blob/a9a7755008a756ca416b304100e44d52bc203ef6/HopfieldNet/aux.lean#L30}{\extlink}
in the update sequence as follows:  
\begin{lstlisting}[numberstyle=\tiny, stepnumber=1, numbersep=5pt, aboveskip=3pt, belowskip=3pt]
def fair (useq : ℕ → U) : Prop := ∀ u n, ∃ m ≥ n, useq m = u
\end{lstlisting}

Similarly, a fixed cyclic update order naturally ensures that every neuron is updated. We define a cyclic update sequence and prove its inherent fairness using the lemma \texttt{cycl\_fair}\href{https://github.com/mkaratarakis/HNBM/blob/a9a7755008a756ca416b304100e44d52bc203ef6/HopfieldNet/aux.lean#L69}{\extlink}
as a sanity check.
\begin{lstlisting}[numberstyle=\tiny, stepnumber=1, numbersep=5pt, aboveskip=3pt, belowskip=3pt]
def cyclic [Fintype U] (useq : ℕ → U) : Prop := 
  (∀ u : U, ∃ i, useq i = u) ∧ 
  (∀ i j, i % card U = j % card U → useq i = useq j)
\end{lstlisting}
We have shown that asynchronous updates, when applied fairly, always lead the Hopfield network to a stable state, forming the basis for its use in associative memory -- a kind of memory that is addressed by its contents.

\subsection{Hopfield Networks as Associative Memory}\label{sec:assoc-memory}

A \emph{pattern} $p = (\mathrm{act}_{u_1}, \ldots, \mathrm{act}_{u_n})^\top \in \{-1,1\}^n$
($n \geq 2$) is a vector of neuron activations -- a possible network state. Unlike classical memory,
which needs the exact state vector, associative memory recovers a stored pattern from partial,
noisy, or corrupted input. Hopfield networks realize this by storing patterns as stable states,
using the Hebbian algorithm~\citeref{hebb} to set the weights and thresholds. Formally, storing $p$
means $S(W p - \boldsymbol{\theta}) = p$, where $W$ is the weight matrix,
$\boldsymbol{\theta} = (\theta_{u_1}, \ldots, \theta_{u_n})$ the threshold vector, and
$S : \mathbb{R}^n \to \{-1,1\}^n$ maps $x \mapsto y$ with $y_i = 1$ if $x_i \geq 0$ and $-1$
otherwise. 

Given a choice of thresholds $\boldsymbol{\theta},$ the goal is to find a suitable weight matrix $W$ such that the stored patterns correspond to stable states. For example, in the case of storing a single pattern $p$ with all thresholds chosen to be zero, it clearly suffices if $W p = c p \text{ with } c \in \mathbb{R}_+.$ 

When storing multiple patterns \( p_1, \ldots, p_m \) with \( m < n \), the weight matrix \( W_i \) is determined for each pattern \( p_i \). The overall weight matrix \( W \) is then the sum of these individual matrices: 
\begin{equation}\label{hebbianmatrix}
W = \sum_{i=1}^{m} W_i = \sum_{i=1}^{m} p_i {p_i}^\top - mI \href{https://github.com/mkaratarakis/HNBM/blob/a9a7755008a756ca416b304100e44d52bc203ef6/HopfieldNet/Quiver/HN/HNquivHebbian.lean#L24}{\extlink},
\end{equation}
where $I$ is the $n\times n$ identity matrix. Since \( p_i^\top p_j = 0 \) for \( i \neq j \) and \( p_i^\top p_i = n \), the inner product of orthogonal vectors is zero, while the inner product of a vector with itself equals its squared length.

The patterns $p_i$ and $p_j$ to be stored can be pairwise orthogonal (that is, if \( p_i^T p_j = 0 \) for \( i \neq j \)) (see \lstinline{patterns_pairwise_orthogonal}), or non-orthogonal (see \lstinline{patterns_pairwise}
\lstinline{_non_orthogonal}). The weight matrix \( W \) for an arbitrary pattern \( p_j \), $j \in \{1, \ldots, m\}$ is given by:

\begin{table}[h]
\centering
\small
\begin{tabular}{@{} p{0.45\textwidth} p{0.50\textwidth} @{}}\hfil \textbf{Orthogonal case}\href{https://github.com/mkaratarakis/HNBM/blob/a9a7755008a756ca416b304100e44d52bc203ef6/HopfieldNet/Quiver/HN/HNquivHebbian.lean#L53}{\extlink} 
& \hfil  \textbf{Non-orthogonal case}\href{https://github.com/mkaratarakis/HNBM/blob/a9a7755008a756ca416b304100e44d52bc203ef6/HopfieldNet/Quiver/HN/HNquivHebbian.lean#L183}{\extlink}
\\
\hfil $\begin{aligned}[t]
W p_j &= \sum_{i=1}^{m} p_i \left( p_i^\top p_j \right) - m p_j \\
      &= (n - m) p_j
\end{aligned}$ & \hfil $ W p_j = (n - m) p_j + \underbrace{
 \sum_{\substack{i = 1 \\ i \ne j}}^{m} p_i (p_i^\top p_j)
 }_{\text{disturbance term}}$ \\
\end{tabular}
\caption{Stability analysis of the Hebbian weight matrix acting on a stored pattern $p_j$.}
\end{table}

In practice, patterns are rarely orthogonal, introducing disturbance terms that affect stability. As $m$ increases, the system becomes more susceptible to these disturbances (as $(n-m)$ shrinks), preventing the network from achieving its theoretical storage capacity.

In Lean, the Hebbian learning rule is the function \lstinline{Hebbian}
\href{https://github.com/mkaratarakis/HNBM/blob/a9a7755008a756ca416b304100e44d52bc203ef6/HopfieldNet/Quiver/HN/HNquivHebbian.lean#L24}{\extlink},
which maps $m$ patterns \lstinline{(ps : Fin m → (HopfieldNetwork R U).State)} to network
parameters \lstinline{Params (HopfieldNetwork R U)} with zero thresholds and the weight matrix
of~\eqrefcol{hebbianmatrix}. As a sanity check, \texttt{HNtest.lean} instantiates a four-neuron Hopfield network with two orthogonal patterns, stores them via Hebbian learning, and demonstrates convergence to a stable stored pattern during the work phase\href{https://github.com/mkaratarakis/HNBM/blob/a9a7755008a756ca416b304100e44d52bc203ef6/HopfieldNet/Quiver/HN/HNtest.lean#L21}{\extlink}.

\section{Stochastic networks} \label{sec:3}
Hopfield networks are described by an energy function: asynchronous updates drive the system to
stable states, and Hebbian learning encodes patterns as such states. This deterministic model is
mathematically clean but restrictive; we generalize to stochastic updates, where a neuron changes
its activation at random with a probability set by the resulting energy change. This links the
energy to a probability distribution and recasts the dynamics as a Markov Chain Monte Carlo (MCMC)
method -- sampling from a distribution by constructing a Markov chain whose long-run behavior
reflects it. In the stochastic setting, \enquote{stability} no longer means a fixed stable state but convergence to a stationary distribution, formalized by reversibility (detailed balance).

A complete MCMC theory has two parts: a measure-theoretic model and a convergence theory. The
model is the \emph{Markov kernel}, a measurable function giving transition probabilities over
arbitrary state spaces; on a finite state space the kernel becomes a stochastic matrix, and
convergence is governed by the Perron--Frobenius theorem and the fundamental theorem of Markov
chains (Thms.~6.2.2/6.2.5, pp.~236--237, \citeref{casella}). We assume familiarity with Markov
kernels, single-site update kernels and basic measure-theoretic probability, following \citeref{casella, kallenberg}, which underlie the \texttt{mathlib} constructions.

On a finite state space (sigma-algebra $\mathcal{A} = \mathcal{P}(\mathcal{X})$), a Markov kernel
is a \emph{stochastic matrix} (pp.~48--52, \citeref{seneta}) with non-negative entries; we adopt the
\emph{column-stochastic} convention, so each column sums to one and the entry $A_{ij}$ is the
probability of transitioning from state $j$ to state $i$.

Intuitively, a single-site update kernel \enquote{resamples} only one coordinate of the state vector while keeping the others fixed, which underlies neuron updates in stochastic finite-state systems such as Boltzmann machines. To generate samples from a distribution $\pi$, we define single-site Gibbs updates (p.~285, \citeref{casella}). 
\begin{definition}[Algorithm A.31, p.~301,~\citeref{casella}]\label{gibbs}
Let $\mathcal{X} = \mathcal{X}_1 \times \dots \times \mathcal{X}_n$ be a finite product state space, and let $\pi$ be a target probability measure on $\mathcal{X}$. For each site $i \in \{1, \dots, n\}$, the \emph{Gibbs update kernel} $K_i$ is a single-site kernel defined by $$K_i(x, A) \;=\; \sum_{x_i' \in \mathcal{X}_i : (x_1,\dots,x_{i-1},x_i',x_{i+1},\dots,x_n) \in A} \pi(x_i' \mid x\setminus{i}),$$
for all measurable sets $A \subseteq \mathcal{X}$, where $x\setminus{i}$ denotes all coordinates of $x$ except the $i$-th. 
\end{definition}
Mathematically, Gibbs sampling is a special case of the Metropolis--Hastings (MH) algorithm~\citeref{hastings}, in which one proposes a move and accepts it with a probability that ensures invariance of the target distribution. In Gibbs sampling, the proposal is the exact conditional distribution of a single coordinate given the others, for which the MH acceptance probability is always 1. The \texttt{TwoState.gibbsUpdate}\href{https://github.com/mkaratarakis/HNBM/blob/a9a7755008a756ca416b304100e44d52bc203ef6/HopfieldNet/Quiver/NeuralNetwork/TwoState.lean#L523}{\extlink} function formalizes the Gibbs kernel for the \lstinline{TwoStateNeuralNetwork} class using \texttt{mathlib}'s \texttt{PMF} type. It instantiates the abstract kernel $K_i$ with the \emph{Boltzmann distribution} below as the target measure $\pi.$ 
\begin{equation}\label{boltzmann}
\pi(\mathbf{act}) = \frac{1}{c} \exp \left( -\frac{E(\mathbf{act})}{kT}\right)
\end{equation}
where $\mathbf{act}$ is the network state, $E$ is the energy of the network, \(c = \sum_{\mathbf{\mathbf{act}}} \exp(-E(\mathbf{\mathbf{act}})/T)\) ensures the probabilities of all possible states equal one, a real number $T >0 $ denoting the temperature of the system, and $k$ is Boltzmann’s constant\footnote{$k \approx 1.38 \cdot 10^{-23} J/K,$ that is, the unit is Joule (energy) divided by Kelvin (temperature).}. We adopt the common simplification in machine learning by replacing $kT$ with $T$, omitting Boltzmann's constant. 

The structure of the random process is defined constructively using Lean's discrete \lstinline{ProbabilityMassFunction} (PMF) monad, consistent with the main choice of structuring our neural network in a linearly ordered field. 

We now make the notion of \emph{reversibility} precise. A Markov kernel $K_i$ is \emph{reversible} with respect to a probability measure $\pi$ if it
satisfies the detailed balance condition (Def.~6.2.1, p.~235, \citeref{casella}). Intuitively, at equilibrium the \enquote{probability flow} between
any two states $s$ and $s'$ is symmetric: the probability of being in $s$ times the transition
probability to $s'$ equals the flow back from $s'$ to $s$. If each single-site kernel $K_i$ is
reversible with respect to $\pi$, so is the \emph{random-scan kernel}
$K_{\mathrm{rand}}(x,A) = \frac{1}{n}\sum_{i=1}^n K_i(x,A)$.

A probability measure $\pi$ is \emph{invariant} for a kernel $K$ if applying $K$ leaves it
unchanged: a system distributed as $\pi$ stays distributed as $\pi$ after any number of steps.
Reversibility implies invariance (\lstinline{IsReversible.invariant}\href{https://github.com/leanprover-community/mathlib4/blob/777913b99d9fe12fbc56a6eb316fca41e4bb4c79/Mathlib/Probability/Kernel/Invariance.lean#L64-L72}{\extlink}, Theorem 6.2.2,
\citeref{casella}): for all $A \subseteq \mathcal{X}$,
$\pi(A) = \int_{\mathcal{X}} K(x,A)\,d\pi(x),$ so $\pi$ is stationary for both the single-site updates $K_i$ and their uniform mixture $K_{\mathrm{rand}}$.

\subsection{Perron--Frobenius theorem}\label{sec:4}
Although \texttt{mathlib} provides the infrastructure for Markov kernels, it lacked a formal
convergence theory for matrices; we supply the first Lean~4 formalization of the Perron--Frobenius
theorem. Reversibility makes $\pi$ invariant but guarantees neither convergence to $\pi$ from an
arbitrary starting state nor its uniqueness -- the stronger property of \emph{ergodicity}. For a
finite state space, a Markov chain is ergodic (Theorem 6.2, \citeref{casella}) iff it is
\emph{irreducible} (every state reaches every other) and \emph{aperiodic} (no trapping in
deterministic cycles). Convergence of finite Markov chains is then formalized via the
Perron--Frobenius theorem for non-negative matrices (Theorem 1.5, \citeref{seneta}). Following
Chapter~1 of \citeref{seneta}, we assume familiarity with real-variable and matrix theory.

A non-negative matrix $A$ is defined as \emph{irreducible} if for every pair of indices $(i, j)$, there exists a power $k$ such that the $(i,j)$-th entry of $A^k$ is positive (\citeref{seneta}, Def. 1.6). 

\begin{theorem}
[\texttt{pft\_irreducible}] \href{https://github.com/mkaratarakis/HNBM/blob/a9a7755008a756ca416b304100e44d52bc203ef6/PF/PerronFrobenius/Irreducible.lean#L398}{\extlink}
Suppose $T$ is an $n \times n$ irreducible non-negative matrix. Then there exists an eigenvalue $r$ such that:
\begin{enumerate}
\renewcommand{\labelenumi}{(\roman{enumi})}
\setlength{\itemsep}{0pt}
    \item$r$ is real, and $r > 0$;
    \item $r$ can be associated with strictly positive left and right eigenvectors
    \href{https://github.com/mkaratarakis/HNBM/blob/a9a7755008a756ca416b304100e44d52bc203ef6/PF/PerronFrobenius/Primitive.lean#L127}{\extlink};
    \item $r \geq |\lambda|$ for any eigenvalue $\lambda$ of $T$
    \href{https://github.com/mkaratarakis/HNBM/blob/a9a7755008a756ca416b304100e44d52bc203ef6/PF/PerronFrobenius/Dominance.lean#L490}{\extlink};
    \item the eigenvectors associated with $r$ are unique up to constant multiples
    \href{https://github.com/mkaratarakis/HNBM/blob/a9a7755008a756ca416b304100e44d52bc203ef6/PF/PerronFrobenius/Multiplicity.lean#L43}{\extlink};
    \item if $0 \leq B \leq T$ and $\beta$ is an eigenvalue of $B$, then $|\beta| \leq r$. Moreover, $|\beta| = r$ implies $B = T.$
    \item $r$ is a simple root of the characteristic equation of $T$\href{https://github.com/mkaratarakis/HNBM/blob/a9a7755008a756ca416b304100e44d52bc203ef6/PF/PerronFrobenius/Multiplicity.lean#L43}{\extlink}.
\end{enumerate}

\end{theorem}

For an irreducible non-negative matrix, the theorem yields a positive real eigenvalue $r$ (the Perron root) with a unique strictly positive eigenvector (up to scaling); for a non-negative column-stochastic matrix (column sums one), $r=1$, and the corresponding right Perron eigenvector, normalized to lie in the standard simplex, is the unique stationary distribution. It further separates aperiodic from periodic matrices: if an irreducible
matrix is aperiodic (or \emph{primitive}), $r$ is the unique eigenvalue of maximum modulus
(p.~14, Theorem 1.1, \citeref{seneta}, \lstinline{pft_primitive}), whereas if it is periodic with
period $d>1$ there are exactly $d$ such eigenvalues (p.~34, Theorem 1.7, \citeref{seneta}).

This is essential for ergodicity: aperiodicity ensures convergence to a unique stationary
distribution, while periodicity yields cyclic behavior.

The Perron--Frobenius theorem is based on studying matrix powers, whose entries encode the number (or total
weight) of fixed-length paths in the associated graph. This correspondence is central to
\emph{irreducibility}: it is defined algebraically by the positivity of matrix powers ($(A^k)_{ij} > 0$),
yet reasoned about combinatorially via strong connectivity of the graph -- the property that every ordered pair
$(u,v)$ is joined by a (possibly empty) directed path. Formally bridging these algebraic and
combinatorial views posed technical challenges, requiring a robust API for path manipulation.

\noindent\textbf{Design decisions}
We require directed edges, ruling out \texttt{SimpleGraph}\href{https://github.com/leanprover-community/mathlib4/blob/777913b99d9fe12fbc56a6eb316fca41e4bb4c79/Mathlib/Combinatorics/SimpleGraph/Basic.lean#L92}{\extlink}, and multiple edges, ruling out \texttt{Digraph}, whose relational adjacency (\secref{sec:2}) abstracts away the identity of individual edges and cannot support path construction and manipulation. \texttt{Graph}\href{https://github.com/leanprover-community/mathlib4/blob/777913b99d9fe12fbc56a6eb316fca41e4bb4c79/Mathlib/Combinatorics/Graph/Basic.lean#L84}{\extlink}, similarly, is undirected. Our solution was to build on \texttt{mathlib}'s \texttt{Quiver} (directed multigraph), which explicitly models paths as sequences of specific, identifiable arrows $(a \rightarrow b).$ We also realized that \texttt{Quiver} was not merely a tool for the proofs, but the correct abstraction for our neural networks themselves (see \secref{sec:2}). We then define a function \lstinline{Matrix.toQuiver}\href{https://github.com/leanprover-community/mathlib4/blob/777913b99d9fe12fbc56a6eb316fca41e4bb4c79/Mathlib/LinearAlgebra/Matrix/Irreducible/Defs.lean#L74-L77}{\extlink}
that maps a matrix $A$ to a \texttt{Quiver}, where an edge $i \to j$ exists if and only if $A_{ij} > 0$. The lemma \lstinline{pow_apply_pos_iff_nonempty_path}\href{https://github.com/leanprover-community/mathlib4/blob/777913b99d9fe12fbc56a6eb316fca41e4bb4c79/Mathlib/LinearAlgebra/Matrix/Irreducible/Defs.lean#L110-L153}{\extlink}
proves that a positive entry in $A^k$ is equivalent to the existence of a path of length $k.$ This allows us to prove essential properties, such as the strict positivity of eigenvectors of irreducible matrices (\lstinline{eigenvector_is_positive_of_irreducible})\href{https://github.com/mkaratarakis/HNBM/blob/a9a7755008a756ca416b304100e44d52bc203ef6/PF/PerronFrobenius/Irreducible.lean#L168}{\extlink}.

The analytic part of the proof (see p.~15 and Theorem 1.1, \citeref{seneta}) centers on the Collatz--Wielandt function: 
\begin{definition}
\label{4.1}
Let $T$ be a non-negative $n \times n$ matrix. The Perron root $r$ of $T$ can be characterized by the Collatz--Wielandt formula which defines the Perron root as
$r = \sup_{\mathbf{x} \ge 0, \mathbf{x} \neq \mathbf{0}} \left( \min_{i \text{ s.t. } x_i > 0} \frac{(T\mathbf{x})_i}{x_i} \right) \href{https://github.com/mkaratarakis/HNBM/blob/a9a7755008a756ca416b304100e44d52bc203ef6/PF/PerronFrobenius/CollatzWielandt.lean#L26}{\extlink}$
\end{definition}

The main challenge is that this function is discontinuous where vector components vanish. We addressed this by proving
it upper-semicontinuous on the standard simplex (non-negative vectors summing to 1), so a
generalized Extreme Value Theorem (\lstinline{exists_isMaxOn_}
\lstinline{of_upperSemicontinuousOn})\href{https://github.com/mkaratarakis/HNBM/blob/a9a7755008a756ca416b304100e44d52bc203ef6/PF/Topology/Compactness/ExtremeValueUSC.lean#L103}{\extlink}
yields a maximum (\lstinline{exists_maximizer})\href{https://github.com/mkaratarakis/HNBM/blob/a9a7755008a756ca416b304100e44d52bc203ef6/PF/PerronFrobenius/CollatzWielandt.lean#L116}{\extlink},
which we show to be the Perron eigenvector (\lstinline{maximizer_satisfies_le_mulVec})\href{https://github.com/mkaratarakis/HNBM/blob/a9a7755008a756ca416b304100e44d52bc203ef6/PF/PerronFrobenius/CollatzWielandt.lean#L531}{\extlink}.
This lemma in turn gives the Perron--Frobenius theorem for stochastic matrices
(pp.~33--34, \citeref{seneta}, \lstinline{exists_positive_eigen}
\lstinline{vector_of_irreducible}\lstinline{_stochastic}\href{https://github.com/mkaratarakis/HNBM/blob/a9a7755008a756ca416b304100e44d52bc203ef6/PF/PerronFrobenius/Stochastic.lean#L24}{\extlink}) -- the main ingredient for the ergodicity of Boltzmann machines.

\subsection{Boltzmann machines}\label{sec:5}

Boltzmann machines are a canonical example of stochastic networks. The energy function is identical to that of Hopfield networks (\eqrefcol{energy}) and ensures that Gibbs updates satisfy detailed balance with respect to the Boltzmann distribution. We describe this construction following (Section 8.6, \citeref{comp}) and \citeref{Geman}. Standard presentations divide neurons into \emph{visible} neurons, which are held fixed to a pattern during the positive phase, and \emph{hidden} neurons, which remain free to update and allow a richer distribution~\citeref{comp,hinton}. 

\noindent\textbf{Design choices.}
We keep the Hopfield-style topology of~\secref{sec:2.4}
($U_{\mathrm{in}} = U_{\mathrm{out}} = U$, $U_{\mathrm{hidden}} = \emptyset$): every neuron is visible and patterns are full activation states. We adopt the binary $\{0,1\}$ convention standard for Boltzmann machines (\lstinline{TwoState.ZeroOne} \href{https://github.com/mkaratarakis/HNBM/blob/a9a7755008a756ca416b304100e44d52bc203ef6/HopfieldNet/Quiver/NeuralNetwork/TwoState.lean#L298}{\extlink}) rather than the $\{-1,+1\}$ bipolar states of Hopfield networks.

Following~\secref{sec:2}, we focus on a single neuron $u$ and how changing its state affects the energy.
For clarity we write updates as probabilities rather than kernels: for a state $\mathbf{act}$ and neuron $u$, $\mathbb{P}(act_u = act_u^{(\text{new})})$ abbreviates the conditional probability $\mathbb{P}(act_u = act_u^{(\text{new})} \mid \mathbf{act}_{\setminus u})$, where $\mathbf{act}_{\setminus u}$ is the activations of all neurons except $u$. We now consider the difference in energy between the two possible states of $u$, $act_u = 0$ and $act_u = 1$, while keeping the states of all other neurons fixed. This difference is 

\begin{equation}\label{eq4bm}
\Delta E_u = E_{\mathrm{act}_u=1} - E_{\mathrm{act}_u=0}
= \theta_u - \sum_{v \in U \setminus \{u\}} w_{uv} \, \text{act}_v
= \theta_u - \mathrm{net}_u.
\end{equation}

Writing the energies in terms of the Boltzmann distribution (\eqrefcol{boltzmann}) yields $\Delta E_u = -T \ln \big(\mathbb{P}(act_u = 1)\big) + T \ln \big(\mathbb{P}(act_u = 0)\big),$ since the probability of $u$ having activation $0$ and the probability of it having activation 1 must sum to 1 (since there are only these two possible states). Solving for the conditional probability $\mathbb{P} (act_u = 1)$ of neuron \(u\) being active gives
\begin{equation}\label{prob}
\mathbb{P}(act_u = 1) = \frac{1}{1 + e^{\Delta E_u /T}},
\end{equation}
a logistic function of the (scaled) energy difference between its active and inactive states. The neuron activation is then set to \(1\) with the probability from \eqrefcol{prob} and to \(0\) otherwise. Concretely, this can be implemented by sampling a random number \(x\) uniformly from the interval \([0,1]\) and updating the neuron according to
\begin{equation}
act_u^{(\text{new})} = 1 \text{ if } x \le \mathbb{P}(act_u = 1),\ \text{else } 0.
\end{equation}
It can be shown that the single-site Gibbs update \lstinline{gibbsUpdateSingleNeuron}
\href{https://github.com/mkaratarakis/HNBM/blob/a9a7755008a756ca416b304100e44d52bc203ef6/HopfieldNet/Quiver/HN/StochasticQuiv.lean#L116}{\extlink} satisfies detailed balance by construction. Let \(\mathbf{act}_{[u \mapsto act_u^{(\text{new})}]}\) denote the state obtained by replacing a randomly selected neuron \(u\)'s activation in \(\mathbf{act}\) with \(act_u^{(\text{new})}\), leaving all others unchanged.
\begin{theorem}[\texttt{detailed\_balance}]
\label{thm:detailed_balance}\href{https://github.com/mkaratarakis/HNBM/blob/a9a7755008a756ca416b304100e44d52bc203ef6/HopfieldNet/Quiver/BM/DetailedBalanceBM.lean#L425}{\extlink} The single-site Gibbs update satisfies detailed balance with respect to the Boltzmann distribution \(\pi\): $$\pi(\mathbf{act}) \, \mathbb{P}\big(act_u = act_u^{(\text{new})}\big) = \pi\big(\mathbf{act}_{[u \mapsto act_u^{(\text{new})}]}\big) \, \mathbb{P}\big(act_u^{(\text{new})} = act_u\big).$$
\end{theorem}

Detailed balance makes $\pi$ stationary for the Gibbs sampler but is local: it does not rule out other stationary distributions. We therefore establish the global property of \emph{ergodicity}: the following theorem shows the sampler has a unique stationary state and converges to it, via the Perron--Frobenius theorem (see~\secref{sec:4})\href{https://github.com/mkaratarakis/HNBM/blob/a9a7755008a756ca416b304100e44d52bc203ef6/PF/PerronFrobenius/Stochastic.lean#L24}{\extlink} for stochastic matrices.

\begin{theorem}[\texttt{randomScan\_ergodicUniqueInvariant}]
\label{thm:ergodic_unique}
\href{https://github.com/mkaratarakis/HNBM/blob/a9a7755008a756ca416b304100e44d52bc203ef6/HopfieldNet/Quiver/BM/Ergodicity.lean#L734-L745}{\extlink}
Let $K$ denote the random-scan Gibbs kernel for a two-state neural network, and let $M_K$ be its corresponding column-stochastic transition matrix. Let $\pi$ be the Boltzmann distribution. The random-scan Gibbs sampler satisfies the following four properties:
\begin{enumerate}
\renewcommand{\labelenumi}{(\roman{enumi})}
\setlength{\itemsep}{0pt}
\item The kernel $K$ is reversible with respect to the Boltzmann distribution $\pi$.
\item Every network state $\mathbf{act}$ has strictly positive self-transition probability,
\(    (M_K)_{\mathbf{act},\mathbf{act}} > 0,
   \)
and hence the Markov chain is aperiodic.
\item The transition matrix $M_K$ is irreducible.
\item There exists a unique probability vector $\mathbf{v}$ in the standard simplex such that
\(    M_K\mathbf{v}=\mathbf{v}.
   \)
\end{enumerate}
\end{theorem}

By (i) and the fact that reversibility implies invariance, established above, $\pi$ is stationary for $K$. Under the finite-state identification of distributions with probability vectors and our column-stochastic convention, the probability vector corresponding to $\pi$ is therefore a fixed point of $M_K$. Since this vector lies in the standard simplex, the uniqueness statement in (iv) implies that it equals $\mathbf{v}$. Hence $\pi$ is the unique stationary distribution of the random-scan Gibbs sampler
\href{https://github.com/mkaratarakis/HNBM/blob/a9a7755008a756ca416b304100e44d52bc203ef6/HopfieldNet/Quiver/BM/Ergodicity.lean#L1052-L1089}{\extlink}.

This construction generalizes Hopfield dynamics: updates are stochastic at high temperature (neurons may change their activation even when energy increases), but as \(T \to 0\) the single-site Gibbs update (\eqrefcol{prob}) becomes deterministic, recovering the asynchronous Hopfield dynamics.
\begin{lemma}[\texttt{gibbs\_update\_tends\_to\_zero\_temp\_limit}]
\label{lem:zero_temp_limit}\href{https://github.com/mkaratarakis/HNBM/blob/a9a7755008a756ca416b304100e44d52bc203ef6/HopfieldNet/Quiver/NeuralNetwork/Zero_temp.lean#L793}{\extlink}
As the inverse temperature goes to infinity (\(1/T \to \infty\)), the probability mass function of the single-site Gibbs update converges pointwise to the deterministic zero-temperature limit kernel.
\end{lemma}
After sufficiently many Gibbs updates, the network reaches \emph{thermal equilibrium}: each state's probability depends only on its energy, not the initial state. It is natural to ask whether $W$ and $\boldsymbol{\theta}$ can be adjusted so the energy function's distribution matches a given \emph{sample of patterns}.

\subsection{Boltzmann learning}\label{sec:5.2}

Where Hebbian learning~(\secref{sec:2.4}) \emph{stores} patterns as stable states in one shot,
with retrieval by convergence (\thmref{hnfairconvergence}), Boltzmann learning instead chooses $W$ and
$\boldsymbol{\theta}$ so that the stationary distribution approximates that of a given sample of
patterns~\citeref{hinton}; a \emph{stochastic associative memory} assigning high probability to
states resembling the stored ones. This succeeds only insofar as the sample is compatible with some Boltzmann distribution. The ergodicity result of~\secref{sec:5} is a prerequisite: it
guarantees the unique stationary distribution that makes the objective well posed.

A standard approach to achieve this task is to measure the difference between two distributions and perform \emph{gradient descent} to minimize that measure (e.g. the Kullback--Leibler divergence~\citeref{kullback1951}). However, we do not formalize this distribution-level objective or its derivation. Instead, the gradient result below is specialized to a single presented pattern $\mathbf{act}^{(0)}$: learning should make the model assign it high probability $\pi(\mathbf{act}^{(0)})$, i.e.\ minimize the \emph{negative log-likelihood}
\begin{equation}
\ell(W,\boldsymbol{\theta}) \;=\; -\log \pi(\mathbf{act}^{(0)})
\;=\; \tfrac{1}{T}\,E(\mathbf{act}^{(0)}) + \log c
\label{eq:negloglik}\href{https://github.com/mkaratarakis/HNBM/blob/a9a7755008a756ca416b304100e44d52bc203ef6/HopfieldNet/BoltzmannLearningQuiver/Learning.lean#L29}{\extlink},
\end{equation}
the model's ``surprise'' at $\mathbf{act}^{(0)}$ (\lstinline{negLogLik}). The partition function $c$
couples all states, so the gradient of $\log c$ is an equilibrium average under $\pi$; this is
what produces the positive-negative phase structure below. We prove the corresponding gradient
identity for this single-pattern objective in \thmref{thm:negLogLik_gradient}.


Standard Boltzmann learning is commonly described in terms of two phases~\citeref{comp,hinton}: in the \emph{positive phase}, statistics are averaged over the training distribution; in the \emph{negative phase}, all neurons are updated freely until equilibrium. In each phase one collects statistics on activations: let $\mathbb{P}_u^+$ be the probability that neuron $u$ is active in the positive phase, $\mathbb{P}_u^-$ the same in the negative phase, and $\mathbb{P}_{uv}^+$, $\mathbb{P}_{uv}^-$ the probabilities that neurons $u$ and $v$ are both active simultaneously in the positive and negative phase respectively. The learning update is~\citeref{hinton}:
\begin{equation}\label{eq:bm-learning-update}
\Delta w_{uv} = \eta\,\bigl(\mathbb{P}_{uv}^+ - \mathbb{P}_{uv}^-\bigr),
\qquad
\Delta \theta_u = -\eta\,\bigl(\mathbb{P}_u^+ - \mathbb{P}_u^-\bigr),
\end{equation}
with learning rate $\eta > 0.$ In our formalization, the positive phase is specialized to a point mass at a single presented pattern $\mathbf{act}^{(0)}$, and \thmref{thm:negLogLik_gradient} proves the corresponding single-pattern gradient result. Intuitively, if neuron $u$ is active more often when a training example is presented than when the network runs freely, its equilibrium probability is too low; decreasing $\theta_u$ makes activation more likely. Similarly, if two neurons are active simultaneously more often under a presented example than under free running, their connection weight should increase, making them more likely to fire together.

The positive-negative update direction~\eqref{eq:bm-learning-update}, \lstinline{learningDirection}\href{https://github.com/mkaratarakis/HNBM/blob/a9a7755008a756ca416b304100e44d52bc203ef6/HopfieldNet/BoltzmannLearningQuiver/Phases.lean#L68}{\extlink},
is packaged as a verified \lstinline{BM.LearningRule}\href{https://github.com/mkaratarakis/HNBM/blob/a9a7755008a756ca416b304100e44d52bc203ef6/HopfieldNet/BoltzmannLearningQuiver/LearningRule.lean#L27}{\extlink}: the positive phase
is the point mass at the fixed pattern (\lstinline{positive}
\lstinline{PhaseAtData}\href{https://github.com/mkaratarakis/HNBM/blob/a9a7755008a756ca416b304100e44d52bc203ef6/HopfieldNet/BoltzmannLearningQuiver/Phases.lean#L56}{\extlink}) and the negative phase is
the Boltzmann equilibrium $\pi$ (\lstinline{negative}
\lstinline{PhaseMeasure}\href{https://github.com/mkaratarakis/HNBM/blob/a9a7755008a756ca416b304100e44d52bc203ef6/HopfieldNet/BoltzmannLearningQuiver/Phases.lean#L52}{\extlink}), which \thmref{thm:ergodic_unique}
identifies as the unique stationary distribution, making the negative-phase averages well defined
(\lstinline{negativePhaseMeasure_is_gibbs_stationary}\href{https://github.com/mkaratarakis/HNBM/blob/a9a7755008a756ca416b304100e44d52bc203ef6/HopfieldNet/BoltzmannLearningQuiver/GibbsStationary.lean#L37}{\extlink}).

For the single pattern
$\mathbf{act}^{(0)}$ the positive phase concentrates on it, so
$\mathbb{P}_{uv}^{+} = \mathrm{act}_u^{(0)}\mathrm{act}_v^{(0)}$ and
$\mathbb{P}_u^{+} = \mathrm{act}_u^{(0)}$, while the negative-phase statistics are the
equilibrium averages $\mathbb{P}_{uv}^{-} = \mathbb{E}_\pi[\mathrm{act}_u\mathrm{act}_v]$ and
$\mathbb{P}_u^{-} = \mathbb{E}_\pi[\mathrm{act}_u]$ (for $\{0,1\}$ activations an activation
probability equals its expectation). With these, \lstinline{learningDirection} is exactly the gradient of $\ell$ from~\eqref{eq:negloglik}:
\begin{theorem}[\texttt{hasGradientAt\allowbreak\_negLogLik\_scaled}]\label{thm:negLogLik_gradient}\href{https://github.com/mkaratarakis/HNBM/blob/a9a7755008a756ca416b304100e44d52bc203ef6/HopfieldNet/BoltzmannLearningQuiver/Learning.lean#L44}{\extlink}
For inverse temperature $\beta = 1/T$, the update of~\eqref{eq:bm-learning-update} is gradient
descent on $\ell$; for weights,
\begin{equation}\label{eq:bm-weight-grad}
\frac{\partial}{\partial w_{uv}} \log \pi(\mathbf{act}^{(0)}) = \beta\,\bigl(\mathrm{act}_u^{(0)}\mathrm{act}_v^{(0)} - \mathbb{E}_\pi[\mathrm{act}_u\,\mathrm{act}_v]\bigr)
\end{equation}
\end{theorem}

Beyond the gradient, we prove the single-pattern loss is convex
(\lstinline{convexOn_negLogLik})\href{https://github.com/mkaratarakis/HNBM/blob/a9a7755008a756ca416b304100e44d52bc203ef6/HopfieldNet/BoltzmannLearningQuiver/Learning.lean#L50}{\extlink} and that a step along \lstinline{learningDirection} strictly decreases it when the data and model statistics differ
(\lstinline{negLogLik_descent_step_scaled}\href{https://github.com/mkaratarakis/HNBM/blob/a9a7755008a756ca416b304100e44d52bc203ef6/HopfieldNet/BoltzmannLearningQuiver/Learning.lean#L55}{\extlink}). Exact negative-phase statistics require long Gibbs runs; for large networks equilibrium is slow, which makes full Boltzmann learning costly~\citeref{comp}. A standard remedy is \emph{contrastive divergence} (CD-$k$), which replaces equilibrium averages by statistics after $k$ full Gibbs sweeps from the presented pattern; we formalize it and prove it convergent to the exact learning direction as $k\to\infty$ (\lstinline{cdLearningDirection_tendsto_learningDirection}\href{https://github.com/mkaratarakis/HNBM/blob/a9a7755008a756ca416b304100e44d52bc203ef6/HopfieldNet/BoltzmannLearningQuiver/CDConvergence.lean#L151}{\extlink}). We omit the construction here, as it is a practical approximation rather than part of the core development, and refer the reader to the repository.

\subsection{Integration with Physlib and Optlib}\label{sec:5.1}
We interpret the network energy as a Hamiltonian $H$ -- exactly the network energy for two-state
networks -- so stochastic updates sample a \emph{canonical ensemble} (a system in thermal equilibrium at $T>0$
favoring lower-energy states; p.~251, \citeref{baran}); the \texttt{IsHamiltonian} typeclass \href{https://github.com/mkaratarakis/HNBM/blob/a9a7755008a756ca416b304100e44d52bc203ef6/HopfieldNet/Quiver/NeuralNetwork/toCanonicalEnsemble.lean#L42}{\extlink} links
\texttt{NeuralNetwork} to \texttt{CanonicalEnsemble}\href{https://github.com/leanprover-community/physlib/blob/ea3c9dd60268886f05c07469b74b38321b975a28/Physlib/StatisticalMechanics/CanonicalEnsemble/Basic.lean#L111}{\extlink}, embedding our networks in \texttt{Physlib}'s
statistical mechanics. The learning side connects to \texttt{Optlib}~\citeref{optlib2025} the same
way: we prove and package the loss and gradient of the single-pattern Gibbs negative log-likelihood
(\thmref{thm:negLogLik_gradient}, \lstinline{convexOn_negLogLik}), with the fixed-step update
\lstinline{gradientDescentFixStep}\href{https://github.com/mkaratarakis/HNBM/blob/a9a7755008a756ca416b304100e44d52bc203ef6/HopfieldNet/BoltzmannLearningQuiver/OptlibBridge.lean#L30}{\extlink}, in \lstinline{OptlibBridge.Objective}. We do not import
\texttt{Optlib}; the bridge records only the expected interface, so a verified multi-step procedure
can later reuse its convergence theorems without reproving gradient descent.

\section{Related Work}\label{sec:6}

Beyond the related work in \secref{sec:1} and \secref{sec:2}, the closest to ours in convergence is
Murphy et al.\ \citeref{perceptron2017}, who formalized and proved convergence of the one-layer perceptron and its
averaged variant in Rocq. Their convergence proof establishes a finite bound on the number of misclassifications made during perceptron training on linearly separable data, whereas our Hopfield convergence proof uses the energy function and the finiteness of the state space to show that the network state stabilizes under asynchronous updates. Otherwise, machine learning formalizations remain sparse: PAC learnability
for decision stumps in Lean \citeref{repara, pac}, the expressiveness of deep learning in
Isabelle/HOL \citeref{expressiveness2018}, AI planning via a SAT encoder \citeref{aiplanning2020},
and, in Rocq, Dvoretzky's convergence theorem, value/policy iteration \citeref{vajjha, certrl},
expectation semantics, and support vector machines \citeref{bhat2013syntactic}.

Formalizations inspired by the sequential perspective include the sequential models in
Isabelle/HOL \citeref{isabelleNN, isabellethesis} and the \texttt{kernel} representation in
Rocq's MLCERT \citeref{mlcert2019}. This approach suits verification goals that closely reflect
real-world implementations, where interoperability with external frameworks matters. Prior
Isabelle/HOL work \citeref{isabelleNN} used dual encodings -- one graph-based for proofs, one for
computation -- enabling formal reasoning alongside on-demand execution in frameworks such as
TensorFlow, and, inspired by it, a Rocq development \citeref{pwa} incorporated external MILP and SMT solvers for
similar objectives. For interoperability, \citeref{mlcert2019} relied on an unverified Python
script to import models, whereas \citeref{isabelleNN} proved correctness for each TensorFlow
import. The sequential model also suits localized computations such as tensor operations or
backpropagation correctness (Certigrad \citeref{certigrad2017}); and in proof assistants enforcing
provable termination, such as Rocq, feedforward networks can be defined inductively and evaluated
by structural recursion, which ensures termination \citeref{mlcert2019}. However, the sequential
layer paradigm struggles with arbitrary topologies, especially the cycles of recurrent networks,
and even general computational graphs (DAGs) often need complex unfolding or fixed-point semantics
to faithfully model true recurrence. Concurrently, TorchLean~\citeref{george2026torchlean} unifies specification, execution, and verification of neural networks in Lean~4; its Hopfield case study builds on the present work~\citeref{ourpaper}, complementing our focus on global recurrent/stochastic dynamics.

Probability theory has been formalized across several proof assistants -- Doob's martingale
convergence in Lean \citeref{doob}, limit theorems and hidden Markov chains in HOL4
\citeref{hol4description, liu}, discrete probability in Mizar \citeref{mizar}, Shannon's theorems
and randomized-concurrency logics in Rocq \citeref{affeldt2012itp, polariscoq}, and
measure-theoretic results including the central limit theorem and the law of large numbers in
Isabelle/HOL \citeref{avigad2017formally, lln2021isabelle, probzoo}. Closest to our Markov-chain
infrastructure is H\"olzl's Isabelle/HOL formalization of Markov kernels via the Giry monad and
the Ionescu--Tulcea extension theorem \citeref{holzl2017markov}; in contrast, we target
finite-state kernels with reversibility and a Perron--Frobenius--based convergence theory aimed at
the ergodicity of the dynamics.

The most extensive prior formalization of Perron--Frobenius, by Thiemann et al.\
\citeref{isabellepf} in Isabelle/HOL, is topological, establishing the Perron eigenvector via
Brouwer's fixed-point theorem. Ours is, to our knowledge, the first analytic formalization, via the
Collatz--Wielandt formula \defref{4.1}: we prove its upper-semicontinuity, identify its maximizer
on the standard simplex with the Perron eigenvector, and give novel proofs of uniqueness and
spectral dominance from the phase alignment of complex eigenvectors, distinct from the topological
method. Where Thiemann et al.\ bridge separate matrix libraries \citeref{HMA, JNF}, we work in
\texttt{mathlib}'s unified linear algebra, graph theory, and analysis, enabling a novel
characterization of irreducibility via quiver strong connectivity (\secref{sec:4}). Finally, their
development targets the full peripheral spectrum and reducible matrices for program analysis,
whereas ours is a foundation for Markov-chain ergodicity and MCMC.

Closer to the natural sciences, a growing body of work formalizes physics in proof assistants. The largest Lean~4 effort, \texttt{Physlib} \citeref{toobysmith2024heplean}, spans electromagnetism,
quantum mechanics, and statistical mechanics, with index notation \citeref{toobysmith2024index}
and Wick's theorem \citeref{toobysmith2025wick}, and even uncovered an error in the published
two-Higgs-doublet-model literature \citeref{toobysmith2026twohiggs}; Lean-QuantumInfo
\citeref{meiburg2024quantuminfo} underpins a machine-checked generalized quantum Stein's lemma
\citeref{meiburg2025stein}. Beyond Lean, chemical thermodynamics and adsorption have been
formalized \citeref{bobbin2024chemphys}, and relativity in Isabelle/HOL~\citeref{stannett2014relativity, schmoetten2022minkowski} and Rocq~\citeref{lu2017specrel}.

\section{Conclusion and future work} \label{sec:7}

This paper presents a multigraph-based framework for neural networks spanning recurrent and feedforward architectures, showing that proof assistants can faithfully capture the interplay of their combinatorial, probabilistic, and thermodynamic aspects, and establishing a verified bridge between statistical mechanics, machine learning, and formal mathematics -- to our knowledge, the first to be mechanized. We outline several directions it opens below.

On the probabilistic side, a general MCMC library would extend our Markov chain formalization.
This requires completing the Perron--Frobenius theorem for general non-negative matrices and the
Fundamental Theorem of Markov Chains in \texttt{mathlib}, providing a foundation for advanced
convergence analysis \citeref{convanalysis} and for verifying sampling algorithms beyond the
single-site Gibbs updates formalized here, such as Metropolis--Hastings \citeref{hastings} and
simulated annealing (p.~142, Example 4.2.4, \citeref{casella}).

On the learning side, we formalize Boltzmann learning on fully visible $\{0,1\}$ networks with a simplified positive phase and single-pattern gradients (\secref{sec:5.2}). Natural extensions include visible/hidden architectures (only hidden neurons equilibrate, visible pattern fixed), a Kullback--Leibler derivation linking phase statistics to gradient descent on the data distribution, and restricted Boltzmann machines. Completing the \texttt{OptlibBridge} (\secref{sec:5.1}) would yield machine-checked multi-step convergence for the verified learning direction; deeper \texttt{Physlib} integration would let us formalize other convergence properties across deterministic and stochastic settings.

On the computational side, our formalization of discrete Hopfield networks is executable but inefficient; still, it can serve as a target specification for verifying a realistic implementation by refinement, in Lean~4 or a more hardware-oriented framework outside of Lean. Making it fully computable would require computable reals such as Rocq's CoRN~\citeref{corn}, alongside Coquelicot~\citeref{coquelicot} for the analysis and Flocq~\citeref{flocq} for floating point; in Lean, Meiburg's computable reals~\citeref{alex} are a promising option.

\subsection*{Acknowledgments}
We thank the anonymous reviewers for their constructive suggestions; Britt Anderson for proposing this project; and Niels van der Weide for his valuable feedback on earlier versions of this manuscript.

\bibliographystyle{plain}
\bibliography{easychair}

\end{document}